\pdfoutput=1
\documentclass[11pt]{article}
\usepackage[final]{acl}
\usepackage{ndnlp}
\usepackage{times}
\usepackage{latexsym}
\usepackage[T1]{fontenc}
\usepackage[utf8]{inputenc}
\usepackage{microtype}
\usepackage{inconsolata}
\usepackage{graphicx}
\usepackage{enumitem}
\setlist{nolistsep}

\usepackage{microtype}
\usepackage{xcolor}  
\usepackage{hyperref}
\definecolor{darkblue}{rgb}{0, 0, 0.5}
\hypersetup{colorlinks=true,citecolor=darkblue, linkcolor=darkblue, urlcolor=darkblue}
\usepackage{booktabs}

\usepackage[T1]{fontenc}
\usepackage[utf8]{inputenc}
\usepackage{adjustbox}
\usepackage{amssymb}
\usepackage{pifont}
\usepackage{multirow}
\usepackage{mathtools}

\usepackage[capitalize,nameinlink,noabbrev]{cleveref}

\let\origpm\pm
\renewcommand{\pm}{\;\origpm}

\title{We're Calling an Intervention: Exploring Fundamental Hurdles in Adapting Language Models to Nonstandard Text} 

\author{Aarohi Srivastava \and David Chiang \\
    Computer Science and Engineering \\
 University of Notre Dame \\
 Notre Dame, IN, USA \\ \texttt{\{asrivas2, dchiang\} @nd.edu}}

\begin{document}
\maketitle

\begin{abstract}
We present a suite of experiments that allow us to understand the underlying challenges of language model adaptation to nonstandard text. We do so by designing \emph{interventions} that approximate core features of user-generated text and their interactions with existing biases of language models. Applying our interventions during language model adaptation to nonstandard text variations, we gain important insights into when such adaptation is successful, as well as the aspects of text variation and noise that are particularly difficult for language models to handle. For instance, on text with character-level variation, out-of-the-box performance improves even with a few additional training examples but approaches a plateau, suggesting that more data is not the solution. In contrast, on text with variation involving new words or meanings, far more data is needed, but it leads to a massive breakthrough in performance. Our findings reveal that existing models lack the necessary infrastructure to handle diverse forms of nonstandard text, guiding the development of more resilient language modeling techniques. We make the code for our interventions, which can be applied to any English text data, publicly available.
\end{abstract}

\section{Introduction}

\begin{figure*}[t]
  \centering
  \includegraphics[width=\linewidth]{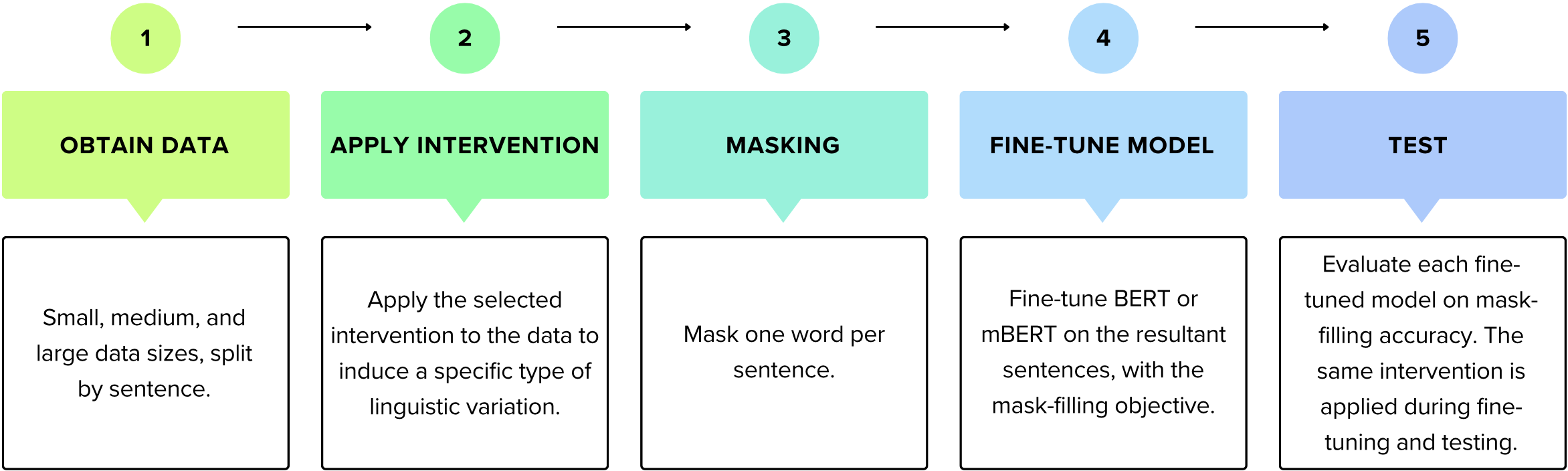}
  \caption{A sketch of our train/test pipeline.}
  \label{fig:diagram}
\end{figure*}

Nonstandard text is all around us. Whether a user adopts a regional dialect, follows different spelling conventions, or uses culturally-specific vocabulary, encountering text variation in most day-to-day NLP use cases is inevitable \citep{blodgett-etal-2016-demographic, huang2020challenges}. Yet, recent work continues to find large gaps in performance between standard and nonstandard text and speech \citep{kantharuban2023quantifying, faisal-etal-2024-dialectbench}, and efforts to reduce this gap are typically case-, variety-, or task-dependent \citep{held2023tada, joshi2024natural}. The distinction between standard and nonstandard text is often contextual. In written English, \emph{going to} is standard, while \emph{gonna} or \emph{gunna} is nonstandard. Likewise, \emph{I am not} (standard) contrasts with \emph{I ain't} (nonstandard). Similarly, \emph{color} (American English) vs. \emph{colour} (British English) reflects orthographic variation. Whether deploying a large-scale system for diverse users or working towards language model personalization, it is important to understand the challenges of adapting a language model to different varieties. Specifically, how can we improve a model's performance in a new domain (in this case, a new variety) by leveraging knowledge from an existing one (i.e., the standard one)?

This is tricky; it is difficult to tease apart the interactions between the complex and intertwined features that comprise linguistic variation, and the black-box nature of language models makes it even more difficult to do so. But we can reach some firmer conclusions if, on the one hand, we can devise ways of controlling different levels of text variation, and, on the other hand, we study the structure of the model (as opposed to its parameter values). We find that the model structure itself induces biases towards the standard variety of a language, and it does so in different ways at different levels of linguistic structure.

Current language models like BERT \citep{devlin2019bert} and GPT \citep{radford2019language} are actually a hybrid of two models: a frequency-based subword tokenizer and a Transformer-based encoder or decoder. The tokenizer determines subword units, and the Transformer embeds the subwords into a vector space where it operates on the vectors. Frequent words are often kept together as a single token, while infrequent words are often broken up into several shorter-length tokens. When both the tokenizer and the Transformer are biased towards standard text, a rigid relationship develops between tokens and their vector representations. Small changes to the token sequence (e.g., resulting from spelling variation) can break the model's ability to understand it properly \citep{kumar2020noisy, soper2021bart, blaschke2023does, srivastava2023bertwich, chai2024tokenization}. See the example below, where $\stackrel{\text{tok}}{\leadsto}$ denotes tokenization:
\begin{center}
  coffee $\stackrel{\text{tok}}{\leadsto}$ coffee \\
  cofee $\stackrel{\text{tok}}{\leadsto}$ co, fe, e \\
\end{center}
While the model would associate the token \emph{coffee} with the appropriate word, one misspelling in the input means the model must figure out that the token sequence \emph{co, fe, e} refers to the same thing as \emph{coffee}, which is certainly not trivial. Moreover, since the subword tokenizer of a pre-trained model cannot be modified without large-scale retraining, such issues can typically only be addressed through model adaptation methods (e.g., fine-tuning). However, because this approach does not resolve the underlying problem, it is crucial to explore the challenges and success cases in adapting language models to different forms of nonstandard text.

To this end, we develop synthetic manipulations that exploit our knowledge of what happens when a Transformer-based language model interacts with nonstandard text. Our experiments isolate data-related factors that can play a role in language model adaptation (e.g., type, amount, and composition of training data), and we assemble a suite of nine interventions to synthetically induce tokenization and embedding disruption, grounded in traits of user-generated text and at different levels of linguistic structure (e.g., orthographic, morphological, lexical), in controlled settings. We make the code for these interventions publicly available.\footnote{\url{https://github.com/aarsri/interventions-linguistic-variation}} 
Our experiments (outlined in \cref{fig:diagram}) evaluate and stress-test BERT’s ability to adapt to various types of nonstandard text under different conditions. Our findings inform important questions about the fundamental hurdles of adapting language models to nonstandard text:
\begin{itemize}
    \item Language models adapt more effectively to lexical variation at the subword and multi-subword levels (e.g., new words, word senses, and meanings) during fine-tuning but struggle with within-subword variation (e.g., character-level changes, unconventional spellings).
    \item When handling lexical variation, greater data availability is vital for successful knowledge transfer. Multilingual models are also more helpful in such cases.
    \item In contrast, for text with character-level variation (within-subword changes), increasing data offers limited benefits. Instead, achieving robustness likely requires alternative solutions. Monolingual models outperform multilingual ones in such cases.
\end{itemize}

\section{Related Work}
\label{sec:relatedwork}
As larger language models with new capabilities emerge, engaging with nonstandard text (e.g., dialects, language varieties, noisy text) -- a hallmark of content generated by today’s diverse user base -- remains a challenge. Language models are not robust to noisy text, plummeting in performance when faced with seemingly simple issues like misspellings, typos, and grammatical errors, even though such issues naturally arise in almost any use case \citep{kumar2020noisy, yin2020robustness, aspillaga2020stress}. These limitations have been documented even in large models. For instance, \citet{pagnoni2024byte} find that Llama 3 and 3.1 perform poorly in robustness tests involving character-level noise, and \citet{chai2024tokenization} highlight the sensitivity of large language models to character-level variations.

Improving model performance on nonstandard text is not trivial. For instance, \citet{faisal-etal-2024-dialectbench} document a persistent performance gap between standard and dialectal text, even after in-variety fine-tuning. Approaches to increasing robustness to linguistic variation, though often successful, tend to be highly dependent on the language variety and task, typically requiring data for each dialect in question \citep{held2023tada} and exposing differences in performance across tasks and varieties \citep{srivastava2023fine}. Given the vast, unpredictable nature of user-generated text, which often involves multiple types of linguistic variation at once, these challenges become even more pronounced. Thus, it is essential to take a step back and develop a deeper understanding of why such adaptation remains difficult and what is needed to facilitate effective learning.

Past work has explored challenges of cross-lingual transfer \citep{philippy2023towards}. For instance, \citet{wu-etal-2023-oolong} investigate three key factors -- embedding space disruption, tokenization changes, and word order changes -- by transforming GLUE datasets to induce each type of shift. They find that while language models can adapt to tokenization and word order changes, they struggle to recover from embedding space disruption (e.g., learning new alignments). Similar conclusions are drawn by \citet{deshpande2022bert} and \citet{jain2022vocab}; \citet{deshpande2022bert} highlight the importance of subword overlap and token embedding alignment for successful knowledge transfer, while \citet{jain2022vocab} suggest leveraging word alignment and dictionary matching techniques. Building on these findings, our experiments examine core features of nonstandard text and explore their interactions with these modeling challenges.

\section{Levels of Text Variation}
\label{variation}

Variation is a natural response to the productive nature of language. It can be observed at all levels of linguistic reasoning \citep{haber1976leaped, geeraerts1994structure}: character-level change (e.g., spelling, abbreviation, orthography), morphological and syntactic variation (i.e., word and sentence structure), lexical and semantic variation (new word senses and meanings), and variation in style and tone, often related to language-external sociopolitical factors. Style and tone variation operate at a higher level of reasoning, integrating features from the first three categories. Viewing user-generated text in terms of these core features elucidates the challenges language models face in adapting to its variability.

When dealing with nonstandard text, we encounter infrequent strings far more often, which inherently challenges existing biases of language models and impacts downstream performance. The tokenizer produces longer sequences of tokens comprised of shorter subwords, a phenomenon called \emph{oversegmentation} \citep{soper2021bart, srivastava2023bertwich}. Shorter subwords can appear in many more contexts and take on different meanings each time, a phenomenon called \emph{subword replacement} \citep{srivastava2023bertwich}. Even in the case of new word senses/meanings, for which the token sequence may not change as much, the model would lack the relevant knowledge. 

Drawing on documented features of user-generated text, we have devised a suite of 9 interventions that operate at different levels of linguistic structure. We apply these interventions to all text used in our experiments. The interventions fall into four categories: character-level change, subword boundary manipulation, morphological variation, and lexical variation. All the interventions require the model to learn to map between standard (seen) and nonstandard (rare or unseen) text for effective knowledge transfer. However, differences in how the interventions affect tokenization can influence the difficulty of learning these mappings during fine-tuning. For this reason, the interventions are designed to cause a fair amount of disruption, with some acting as stress tests. All of our data and experiments are in English, but the algorithmic nature of many of our interventions can be extended to other languages.

\subsection{Character-Level Change}
Character-level change can arise for several reasons, including phonological influences (e.g., accent, sound change), new methods of writing (e.g., social media), and typological errors \citep{condorelli2023cambridge}. We include two interventions under this category; the first is a reflection of phonological variation, and the second is more of a stress test.
\begin{enumerate}
  \item \textbf{IPA}: Letters corresponding to consonants on the IPA chart that form a minimal feature pair (voiced vs. unvoiced) are swapped. For example, \emph{p} and \emph{b}, which differ only in the voicing feature but share the same place and manner of articulation, are swapped. Changes to consonants are a hallmark of modern English orthographic variation, particularly observed on social media \citep{eisenstein2015systematic, ilbury2020sassy}. For instance, such variations are prevalent in corpora of noisy text like MultiLexNorm \citep{van2021multilexnorm}; examples include dese/these, dey/they, smyle/smile, and sadest/saddest.
  \begin{center}
    boots $\stackrel{\text{IPA}}{\leadsto}$ poodz $\stackrel{\text{tok}}{\leadsto}$ p, ood, z
  \end{center}
  \item \textbf{Shift}: A Caesar cipher is applied to the letters of the alphabet: $a \rightarrow b, b \rightarrow c, ..., z \rightarrow a$. This intervention is the most extreme form of orthographic change, in which every alphabetic symbol is renamed. It roughly approximates situations where a language variety uses a different alphabet than the standard variety or where social media trends using symbols that resemble letters. \textbf{Shift} serves as a stress test, as well as a means of comparison to milder interventions like \textbf{IPA}.
  \begin{center}
    boots $\stackrel{\text{Shift}}{\leadsto}$ cpput $\stackrel{\text{tok}}{\leadsto}$ c, pp, ut
  \end{center}
\end{enumerate}
While orthographic changes like the ones above would ordinarily only require us to learn to map one character to another (or one spelling of a word to another, as in texting abbreviations), the structure imposed by the model makes this task more complex. As exemplified above, adapting to this category of variation would require the model to learn a one-to-many mapping from a single token to a list of tokens.

\subsection{Subword Boundaries}
As demonstrated above, linguistic variation often results in changes to subword token boundaries, which triggers a domino effect and ultimately results in lower quality contextual embeddings assigned to nonstandard text by the model \citep{kumar2020noisy, soper2021bart}. Because of this, we include three interventions that overtly manipulate subword boundaries, two of which preserve the original spelling of the word and only split tokens. Successful adaptation would once again involve learning a one-to-many mapping, but with more of the surface level appearance (i.e., spelling) intact. Examples like ``ammmazing'' (amazing) in MultiLexNorm \citep{van2021multilexnorm} resemble this situation.

\begin{enumerate}
  \item \textbf{Reg}: Subword regularization using the MaxMatch Dropout \citep{hiraoka2022maxmatch} method for BERT's WordPiece algorithm is applied with a dropout of 0.5. This means 50\% of the subword vocabulary is randomly dropped with each tokenization call.
  \begin{center}
    boots $\stackrel{\text{tok}}{\leadsto}$ boots 
    \smallskip
    
    boots $\stackrel{\text{Reg}}{\leadsto}$ boot, s
  \end{center}
  \item \textbf{Char}: Subword regularization using the MaxMatch Dropout \citep{hiraoka2022maxmatch} method for WordPiece is applied with a dropout of 1, meaning each letter of a word is its own token.
  \begin{center}
    boots $\stackrel{\text{Char}}{\leadsto}$ b, o, o, t, s
  \end{center}
  \item \textbf{Pig}: Words are converted to Pig Latin, in which the word-initial consonant(s) is moved to the end, and a suffix (\emph{ay} or \emph{yay}) is added. For example, \emph{pig latin} becomes \emph{igpay atinlay}.
  \begin{center}
    boots $\stackrel{\text{Pig}}{\leadsto}$ ootsbay $\stackrel{\text{tok}}{\leadsto}$ o, ots, bay
  \end{center}
\end{enumerate}

\subsection{Morphological Variation}
In English, morphological variation can occur in inflectional or derivational affixes \citep{neef2009morphological, zanuttini2014micro}. Inflectional endings are morphemes with grammatical functions that typically change grammatical features like part of speech and number (e.g., plural \emph{-s}). This is a type of morphosyntactic variation, which relates to grammatical acceptability and is often highly stigmatized -- some utterances may be grammatical in one variety and not in the standard, or vice versa. In contrast, derivational affixes are used to change the meaning of a word, which may or may not also change its part of speech. Unlike inflectional endings, which are a functional category, derivational affixes have much more room for variation. 

\begin{enumerate}
  \item \textbf{$-$End}: Using MorphyNet \citep{batsuren2021morphynet}, inflectional endings are dropped from words that have them.
  \begin{center}
    boots $\stackrel{\text{$-$End}}{\leadsto}$ boot $\stackrel{\text{tok}}{\leadsto}$ boot
  \end{center}
  \item \textbf{Affix}: Using MorphyNet \citep{batsuren2021morphynet}, derivational prefixes and suffixes are mapped cyclically. For instance, \emph{non-} becomes \emph{ab-}, \emph{ab-} becomes \emph{pre-}, etc. \emph{Nonsense} is now \emph{absense} (not ``absence''), \emph{absence} is now \emph{presence}, and so on. In this way, we tinker with part of a word but change its whole meaning. 
  \begin{center}
    nonsense $\stackrel{\text{Affix}}{\leadsto}$ absense $\stackrel{\text{tok}}{\leadsto}$ a, bs, ense
  \end{center} 
\end{enumerate}
Because the lemma is preserved in these cases, learning the appropriate knowledge will require a mix of recovering token mappings and transferring existing knowledge about the meaning of the word from the lemma.

There are several examples of morphosyntactic variation found in user-generated text \citep{zanuttini2014micro}. A common feature in MultiLexNorm \citep{van2021multilexnorm} seems to be informal contractions (e.g., ``imma,'' ``finna,'' ``tryna,'' ``hella''). Because these forms have lower preservation of the lemma, the model may also see them as new words, bringing us to the final category: lexical variation.

\subsection{Lexical Variation}
Word-level variation can be just that -- a new word introduced to refer to a specific, perhaps novel, referent, or to be used in a new context. But lexical variation can often introduce semantic variation; the typical process being that a word's senses expand and eventually shift to the new meaning, through specialization, generalization, or subjectification \citep{kakharova2021nature, geeraerts2024lexical}. Such occurrences are extremely common on social media and in colloquial settings. 
\begin{enumerate}
  \item \textbf{Hyp}: Exemplifying specialization, through which a word's range of reference narrows, words with hyponyms found in WordNet \citep{wordnet} are replaced by their hyponym.
  \begin{center}
    boot $\stackrel{\text{Hyp}}{\leadsto}$ buskin $\stackrel{\text{tok}}{\leadsto}$ bus, kin
  \end{center}
  \item \textbf{Ant}: Exemplifying subjectification, through which a word's meaning becomes more positive (ameliorization) or negative (pejorization), words with antonyms found in WordNet \citep{wordnet} are replaced by their antonym.
  \begin{center}
    nice $\stackrel{\text{Ant}}{\leadsto}$ nasty
  \end{center}
\end{enumerate}
With the lexical interventions, the model must learn to map between contextual meanings of seen words. Variation is at the subword or multi-subword level.

\subsection{Interventions in Action}
Through these four categories of interventions, we are able to examine a range of underlying factors in modeling and adaptation capability for nonstandard text, from tokenization-specific issues (character-level change and subword boundaries), to token-embedding relationships (morphological variation), to contextual representation shift (lexical variation). Moreover, within each category, we include milder, more realistic interventions (e.g., \textbf{IPA}) and stress tests (e.g., \textbf{Shift}). Our experiments reveal whether a model has a chance at recovering each type of mapping during fine-tuning, and under which data-related conditions.

\section{Data Preparation}
\label{sec:data_prep}
To compare the model's ability to adapt to each synthetic variety, we fine-tune and test it with the mask-filling objective on text with the appropriate intervention applied. We provide a sketch of our train/test pipeline in \cref{fig:diagram}. All of our data is sourced from Wikicorpus \citep{reese2010wikicorpus} to reduce external effects of choice of data. We reserve half the Wikicorpus articles for fine-tuning and half for testing. These are separated into sentences using the sentence tokenizer from NLTK \citep{loper2002nltk}. In our experiments, a \emph{word} is any string that satisfies Python's isalpha function and is an element of NLTK word tokenizer's output. Sentences with 0 or 1 words are eliminated. 

\begin{table}\centering\small
  \begin{tabular}{@{}c|r|cccc@{}}
\toprule
\textbf{Size} & Sentences & \multicolumn{4}{c}{Word Count Quartiles} \\
& & Average & $P_{25}$ & $P_{50}$ & $P_{75}$ \\ \midrule
\textbf{S}  & 264     & 19.9      & 10       & 18       & 26       \\
\textbf{M}  & 2641    & 18.5      & 10       & 16       & 24       \\
\textbf{L}  & 26415    & 18.4      & 10       & 16       & 24       \\ \bottomrule
\end{tabular}
\caption{\label{data} Sentence and word count statistics ($25^{th}$, $50^{th}$, and $75^{th}$ percentiles) for each data split.}
\end{table}

When fine-tuning, we vary the amount of data used. Each split is a fixed set of sentences sampled from the fine-tuning articles. The number of sentences in each split covers three orders of magnitude. We report statistics in \cref{data} on sentence length (measured by number of \emph{words}) for each split to demonstrate that they are comparable in this regard. In addition to varying the data size and intervention, we also vary the composition of the fine-tuning data. It could be \emph{mixed}, meaning the intervention is only applied to half the sentences, or \emph{full}, meaning the intervention is applied to all sentences.

Fine-tuning typically focuses on adapting a model to a specific task, making it challenging to simultaneously adapt the model to a new language variety. Task-specific factors can also complicate this process, especially in experiments involving disrupted text. For example, some tasks, like intent classification, rely on identifying key words in the input, while others, like sentiment analysis, require an overall understanding of the sentence, and more linguistically complex tasks, such as linguistic acceptability, demand deeper grammatical reasoning. These variations create a complex interaction between handling perturbed text and meeting the requirements of each task, potentially introducing bias into the experiment. To avoid these complications, we select mask-filling as our fine-tuning task. Mask-filling aligns with the models’ pre-training objective, allowing fine-tuning resources to focus exclusively on adapting to the new language variety without interference from task-specific factors.

Our masking policy during fine-tuning is as follows. One \emph{word} per sentence is randomly selected to be masked. We practice whole-word masking, meaning the model could be asked to fill one or more consecutive mask tokens, given the number of subword tokens that comprise the masked word. Depending on the word to be masked, it is possible the intervention does not actually change that word. While this is natural for fine-tuning, to make sure the comparison during testing is fair, we mandate that the word masked at test time is actually modified by all nine interventions. This filtering yields a test set of 931 sentences sampled from the testing articles. For each sentence, the same word is masked in all interventions/tests for consistency.

\section{Experiments}

There are four axes of variation in our experiments: \emph{amount} of fine-tuning data (small, medium, large), \emph{composition} of fine-tuning data (50\% (mixed) or 100\% (full) of sentences are noised), \emph{intervention} to be applied (9 total), and \emph{multilinguality} (monolingual or multilingual pre-trained model). We use BERT-base-cased\footnote{\url{https://huggingface.co/bert-base-cased}} (BERT) as the monolingual English model, and BERT-base-multilingual-cased\footnote{\url{https://huggingface.co/bert-base-multilingual-cased}} (mBERT) as the multilingual model. We do not include character-level models like CharacterBERT \citep{el2020characterbert} and CANINE \citep{clark2022canine} in our experiments; unlike BERT and mBERT, they cannot be used out-of-the-box for mask-filling and would not yield comparable results. For larger models, it is often unclear whether success stems from pre-training exposure, parameter-based knowledge retention, or true adaptation (our focus). Using BERT, we minimize such confounds.

For each possible combination of the four axes of variation, a pre-trained model is fine-tuned on the corresponding data with the masked language modeling objective. As described in \cref{sec:data_prep}, one word per sentence is masked (whole word masking), and the fine-tuning objective is to fill the masked token(s) with the tokens of the original word.

We use Low-Rank Adaptation (LoRA, \citet{hu2021lora}) for parameter-efficient fine-tuning, which adapts the attention weights of each Transformer encoder layer and freezes all other parameters. We follow \citeauthor{hu2021lora}'s guidelines for hyperparameter choice, using the AdamW optimizer \citep{loshchilov2018decoupled} with a linear scheduler, LoRA rank of 8, and LoRA scaling factor $\alpha$ of 8. We use learning rate $7 \cdot 10^{-4}$ for the small data amount and $5 \cdot 10^{-4}$ for the medium and large data amounts. On an NVIDIA A10 Tensor Core GPU, fine-tuning takes 12 seconds/epoch for the small data amount, 2 minutes/epoch for the medium data amount, and 18 minutes/epoch for the large data amount. LoRA was chosen over standard fine-tuning for two key reasons: (1) standard fine-tuning is highly susceptible to distribution shift issues, and (2) LoRA provides greater control over where learning occurs (in encoder parameters associated with attention).

We measure performance with three metrics: from most to least strict, exact match, 1-best, and 5-best accuracy. Exact match accuracy measures for how many masked words each token of the word is predicted correctly, divided by the test set size. 1-best accuracy measures the total number of masked tokens filled correctly (by the top probability prediction) divided by the total number of masked tokens in the test set. Similarly, 5-best accuracy measures the total number of masked tokens whose top five predictions include the correct answer, divided by the total number of masked tokens in the test set. The 1-best accuracy metric provides the best summary of the results \cref{full-1best}; we include results using the other two metrics in Tables~\ref{full-em} and \ref{full-5best}.

\section{Results}
The main 1-best results for our experiments are found in \cref{full-1best}, and the normalized scores are reported in \cref{relative}. Additional results using the other metrics are included in Tables~\ref{mixed-1best}-\ref{full-5best}. We reiterate the categories of interventions (see \cref{variation}) below:
\begin{itemize}
  \item Character-Level Change: \textbf{IPA}, \textbf{Shift}
  \item Subword Boundary Variation: \textbf{Reg}, \textbf{Char}, \textbf{Pig}
  \item Morphological Variation: \textbf{$-$End}, \textbf{Affix}
  \item Lexical Variation: \textbf{Hyp}, \textbf{Ant}
\end{itemize}

\subsection{Baselines: We need an intervention!}
We include a baseline row for each model (data amount 0), in which the pre-trained model, without any additional fine-tuning, is tested on each intervention. We also include a baseline intervention \emph{None}, in which the models are evaluated on the original text without any intervention applied. 

As expected, monolingual BERT is better out-of-the-box at the mask-filling task on English Wikipedia text. Fine-tuning without any intervention results in overfitting -- an expected outcome, as the model has already been trained on the same task with similar data. Thus, we also include relative performance -- each baseline score in \cref{full-1best} is normalized by the corresponding \emph{None} performance and expressed as a percentage in \cref{relative}. Out-of-the-box performance (data amount 0) is extremely low across the board, demonstrating that our experiments will provide clear findings as BERT cannot already solve these tasks.

\begin{table*}[t]\centering\small
\begin{tabular}{@{}cc|rrrrrrrrrr@{}}
\toprule
        &        & \multicolumn{10}{c}{\textbf{Interventions}}                                                                                                                                                                               \\ \midrule
\textbf{Model} & \textbf{Data} & \multicolumn{1}{c}{\textbf{None}} & \multicolumn{1}{c}{\textbf{IPA}} & \multicolumn{1}{c}{\textbf{Shift}} & \multicolumn{1}{c}{\textbf{Reg}} & \multicolumn{1}{c}{\textbf{Char}} & \multicolumn{1}{c}{\textbf{Pig}} & \multicolumn{1}{c}{\textbf{$-$End}} & \multicolumn{1}{c}{\textbf{Affix}} & \multicolumn{1}{c}{\textbf{Hyp}} & \multicolumn{1}{c}{\textbf{Ant}} \\ \midrule
BERT      & 0       & 58.8               & 2.6               & 2.3                & 7.0               & 5.4                & 9.3               & 1.5                & 0.1                & 0.1               & 0.2               \\
        & S       & 52.4               & 3.4               & 5.6                & 9.0               & 12.2               & 15.3               & 32.3                & 0.3                & 2.0               & 2.6               \\
        & M       & 48.6               & 15.3               & 8.7                & 12.7               & 16.1               & 14.8               & 28.9                & 14.5                & 9.5               & 19.1               \\
        & L       & 47.7               & 18.8               & 9.5                & 11.6               & 26.5               & 23.1               & 37.6                & 35.7                & 29.6               & 29.6               \\
 \midrule mBERT     & 0       & 42.0               & 2.6               & 2.2                & 5.1               & 6.4                & 13.8               & 2.6                & 1.7                & 1.7               & 1.1               \\
        & S       & 38.8               & 10.4               & 4.4                & 4.7               & 11.6               & 13.6               & 23.1                & 1.7                & 5.2               & 7.1               \\
        & M       & 31.6               & 12.1               & 3.7                & 12.1               & 20.5               & 16.7               & 30.5                & 18.8                & 17.6               & 19.7               \\
        & L       & 34.4               & 13.2               & 5.2                & 11.1               & 29.7               & 21.3               & 30.0                & 41.7                & 28.9               & 33.6               \\ \bottomrule
\end{tabular}
\caption{\label{full-1best} 1-best accuracy results (single run) for all experiments with the full data composition using the base version of the model. Data amount $0$ denotes the out-of-the-box baseline performance compared to fine-tuning with the small (S), medium (M), or large (L) data sizes.}
\end{table*}

\begin{table*}[t]\centering\small
\begin{tabular}{@{}cc|rrrrrrrrrr@{}}
\toprule
        &        & \multicolumn{10}{c}{\textbf{Interventions}}                                                                                                                                                                               \\ \midrule
\textbf{Model} & \textbf{Data} & \multicolumn{1}{c}{\textbf{None}} & \multicolumn{1}{c}{\textbf{IPA}} & \multicolumn{1}{c}{\textbf{Shift}} & \multicolumn{1}{c}{\textbf{Reg}} & \multicolumn{1}{c}{\textbf{Char}} & \multicolumn{1}{c}{\textbf{Pig}} & \multicolumn{1}{c}{\textbf{$-$End}} & \multicolumn{1}{c}{\textbf{Affix}} & \multicolumn{1}{c}{\textbf{Hyp}} & \multicolumn{1}{c}{\textbf{Ant}} \\ \midrule
BERT      & 0       & 100.0               & 4.4               & 3.8                & 11.9               & 9.2                & 15.7               & 2.6                & 0.2                & 0.1               & 0.3               \\
        & S       & 100.0               & 6.5               & 10.8                & 17.2               & 23.3               & 29.2               & 61.7                & 0.5                & 3.8               & 4.9               \\
        & M       & 100.0               & 31.5               & 18.0                & 26.1               & 33.1               & 30.5               & 59.4                & 29.8                & 19.6               & 39.3               \\
        & L       & 100.0               & 39.4               & 19.9                & 24.4               & 55.4               & 48.3               & 78.8                & 74.9                & 62.0               & 62.1               \\ \midrule
mBERT     & 0       & 100.0               & 6.1               & 5.3                & 12.1               & 15.1               & 32.8               & 6.2                & 4.1                & 4.0               & 2.5               \\
        & S       & 100.0               & 26.9               & 11.4                & 12.1               & 30.0               & 35.2               & 59.6                & 4.3                & 13.3               & 18.3               \\
        & M       & 100.0               & 38.4               & 11.6                & 38.1               & 64.9               & 52.9               & 96.4                & 59.5                & 55.5               & 62.4               \\
        & L       & 100.0               & 38.4               & 15.0                & 32.2               & 86.2               & 61.8               & 87.0                & 120.9               & 83.9               & 97.6               \\ \bottomrule
\end{tabular}
\caption{\label{relative} Relative performance (single run) normalized by the baseline (\emph{None}), as percentages using the base version of the model. Data amount $0$ denotes the out-of-the-box baseline performance compared to fine-tuning with the small (S), medium (M), or large (L) data sizes.}
\end{table*}

\subsection{Mixed vs. Full Composition: What is needed to learn new mappings?}

The knowledge transfer that takes place when adapting a model to a new variety of the language is akin to learning how to map elements of the standard variety to the new one. One of the varied parameters of our experiments is whether the fine-tuning data is mixed (intervention is only applied to 50\% of sentences) or full (intervention is applied to all sentences). Sometimes, we might expect model learning to benefit from seeing both standard and nonstandard versions of text during fine-tuning, while other times, this only makes learning the appropriate patterns more difficult. 

Our results indicate that the latter holds true when dealing with linguistic variation; average performance (not including baseline scores) is about 4.2 points, or 36\%, higher, with the full composition (\cref{full-1best}) rather than the mixed composition (\cref{mixed-1best}). Performance with full composition is also higher when comparing averages for each intervention category, so the mixed composition does not provide an advantage for any type of variation tested. Because of this difference in performance, we refer only to the full composition results when discussing and analyzing the remaining factors below. 

\subsection{Data Size: When is more data needed?}
Across our experiments, particularly looking at the normalized performance relative to the baseline intervention scores (\cref{relative}), more fine-tuning data helps. At the same time, the utility of adding more fine-tuning data differs depending on the type of mapping needed for a class of interventions. 

When it comes to the orthographic interventions (\textbf{IPA}, \textbf{Shift}) and \textbf{$-$End}, the input sequence is affected by tokenization-related issues (i.e., oversegmentation and subword replacement), and the model must learn one-to-many mappings between tokens during fine-tuning for successful knowledge transfer. In this case, simply fine-tuning, even on the small or medium data amounts, results in a big improvement, but further improvements are much smaller. While fine-tuning helps, it is not sufficient to fully recover the mapping due to the model's inability to draw on sub-atomic knowledge (i.e., within-subword character-level information).

Subword boundary manipulation still involves the issues of oversegmentation, but the closeness in spelling may result in a stronger association by the model between the contextual representations of pre- and post-intervention versions of the same words. When adapting to the synthetic varieties with subword boundary manipulation, performance improves gradually as more data is added.

Strikingly, there is an apparent breakthrough effect when fine-tuning with the largest data size for the lexical interventions (\textbf{Hyp}, \textbf{Ant}) and \textbf{Affix}. While the performance attained is comparatively low for the small and medium data sizes, there is a massive jump when the large data size is used for fine-tuning. This is most clearly observed in the exact-match performance (\cref{full-em}) for BERT and the relative performance for mBERT, which nears or exceeds 100\% in these three tasks. These tasks require relearning the usage of words in new contexts. Mapping word-level information (i.e., new spellings) as well as contextual meaning evidently requires much more data, but when the data requirement is satisfied, the model is capable of recovering (multi-) subword level information. 

\subsection{Monolingual vs. Multilingual: Which type of knowledge is more helpful?}
For the most part, the average performance (not including baseline performance) is extremely close between BERT and mBERT. Notably, mBERT provides a substantial advantage for adapting to variation in meaning (\textbf{Affix}, \textbf{Hyp}, and \textbf{Ant}), providing a 20\% boost in absolute scores (\cref{full-1best}) and a 74\% relative improvement in normalized scores (\cref{relative}). Because mBERT is trained on several languages, it is likely not as rigid in terms of relating words to specific meanings/contexts, providing an advantage here.

\subsection{Implications for User-Generated Text}
Our interventions provide insights into how language models adapt to the diverse noise patterns in user-generated text. We examine cases where token sequences are disrupted but meaning is preserved (e.g., character-level changes) and those where meanings shift with minimal impact on tokenization (e.g., lexical variation), as well as intermediate cases like subword boundary manipulation.

Our findings show that models struggle to process variations within subwords. Given the prevalence of character-level changes in user-generated text, particularly on social media, fine-tuning on additional data alone is insufficient for robust adaptation. This poses challenges for NLP applications in these domains, especially when users prefer to retain their stylistic choices rather than conform to standardized text. At the same time, we find that with sufficient data, models can learn certain morphological and lexical variations, making it possible to adapt to new words, slang, and evolving usage patterns. However, given the diversity of user-generated content, some phenomena may be too sparse for effective learning, highlighting the need for more targeted approaches beyond data scaling.

\section{Conclusion}

We introduce a suite of interventions that synthetically modify English text to analyze interactions between features of nonstandard and user-generated text with underlying biases of language models. Our experiments isolate data-related factors that can contribute to language model adaptation, revealing critical insights into the limitations of adapting language models to nonstandard text. We explore several cases, ranging from character-level changes that over-segment the token sequence to lexical variations that alter contextual representations. Some interventions require learning one-to-many mappings within subwords, while others demand associations across multiple subwords.

Our findings highlight key adaptation challenges. BERT-like models struggle to adapt to character-level changes, even with additional data, but can successfully handle lexical shifts and new word senses with enough exposure to relevant fine-tuning data. Notably, interventions involving affix changes and antonym substitutions achieve performance comparable to or exceeding mBERT baselines. This suggests that while models can effectively learn word-to-word mappings, structural constraints hinder their ability to process finer-grained variations within subwords or at the character level. These limitations arise from how tokens are segmented and represented within the model, restricting its ability to capture within-subword (sub-atomic) variations.

Ultimately, while current models adjust well to meaning-related variations given enough data, they struggle with fine-grained structural disruptions, where constituents are smaller than a subword (e.g., character-level). Despite the prevalence of such variation (e.g., social media), current language models lack the capability to facilitate the required flexibility between tokens and embeddings for straightforward adaptation methods like fine-tuning, unless the relevant knowledge is already incorporated during pre-training. This underscores the need for more flexible tokenization and modeling approaches, especially for handling the complexities of user-generated text and enabling the model to effectively capture within-subword (e.g., character-level) information.

\section*{Limitations}
While our experiments explored numerous possibilities within the scope of our study, they are certainly not exhaustive. We recognized the vast array of potential variations and interventions that could be considered and aimed to curate a feasible selection that still offered a diverse representation of linguistic challenges and adaptation scenarios. Beyond the selection of interventions to design, it would be valuable to expand the scope of experiments in the other dimensions, as well: languages, data sizes, and models. 

For instance, our interventions and experiments were developed and executed in English. While narrowing down the language dimension allowed us to develop the highly controlled experiments needed for this study, it also means insights are missing for languages in which the character-, subword-, and word-level paradigm used in this work may not apply in the same way. Examples include languages with richer morphology (e.g., Turkic languages) and languages for which characters correspond to syllables, words, or concepts (e.g., Sino-Tibetan languages). Similarly, while the code for the interventions can largely be extended to other Indo-European languages, it would require more modification before it could be used for other language families.

Furthermore, while our choices for the small, medium, and large data sizes are informed by runtime and typical fine-tuning data sizes in NLP work, some of our results point to trends as data size increases. As a result, it would be valuable to extend these tables to see an even bigger picture as data size continues to increase. In addition, while we used LoRA for model adaptation in this study, there are other approaches that could be explored to better understand the nuances of parameter-efficient fine-tuning.

Finally, our work focuses on the widely-used BERT, which is an Encoder-only language model of relatively small size. While the small size is beneficial for our study, as it assures us that the model has not already seen the intervention tasks during pre-training, diversity in the type of model (e.g., encoder-decoder, decoder-only, LLMs) can help paint a bigger picture for our results and their implications in a wider range of settings. 

\bibliography{custom}

\appendix
\section{Appendix}
\begin{table*}[h]\centering\small
\begin{tabular}{@{}cc|rrrrrrrrrr@{}}
\toprule
\multicolumn{1}{c}{}        & \multicolumn{1}{c|}{}       & \multicolumn{10}{c}{\textbf{Interventions}}                                                                                                                                                                                     \\ \midrule
\multicolumn{1}{c}{\textbf{Model}} & \multicolumn{1}{c|}{\textbf{Data}} & \multicolumn{1}{c}{\textbf{None}} & \multicolumn{1}{c}{\textbf{IPA}} & \multicolumn{1}{c}{\textbf{Shift}} & \multicolumn{1}{c}{\textbf{Reg}} & \multicolumn{1}{c}{\textbf{Char}} & \multicolumn{1}{c}{\textbf{Pig}} & \multicolumn{1}{c}{\textbf{$-$End}} & \multicolumn{1}{c}{\textbf{Affix}} & \multicolumn{1}{c}{\textbf{Hyp}} & \multicolumn{1}{c}{\textbf{Ant}} \\ \midrule
BERT                & 0                 & 58.8               & 2.6               & 2.3                 & 6.9               & 5.4                & 9.3               & 1.5                & 0.1                & 0.1                 & 0.2                 \\
                  & S                 & 52.4               & 2.4               & 5.9                 & 11.0               & 11.7               & 17.4               & 18.5                & 0.8                & 0.6                 & 0.8                 \\
                  & M                 & 48.6               & 7.5               & 8.3                 & 12.5               & 13.1               & 16.3               & 26.7                & 10.6                & 8.3                 & 3.0                 \\
                  & L                 & 47.7               & 14.4               & 8.3                 & 12.0               & 19.9               & 20.3               & 15.1                & 25.5                & 17.0                 & 19.7                 \\
\midrule mBERT               & 0                 & 42.0               & 2.6               & 2.2                 & 5.1               & 6.4                & 13.8               & 2.6                & 1.7                & 1.7                 & 1.1                 \\
                  & S                 & 38.8               & 10.5               & 1.5                 & 7.4               & 12.3               & 10.1               & 5.2                & 4.2                & 1.3                 & 6.8                 \\
                  & M                 & 31.6               & 9.3               & 4.8                 & 12.8               & 14.1               & 16.2               & 7.6                & 24.0                & 9.7                 & 8.5                 \\
                  & L                 & 34.4               & 14.7               & 6.8                 & 13.0               & 16.4               & 16.9               & 11.9                & 29.5                & 25.3                 & 22.8                 \\ \bottomrule
\end{tabular}
\caption{\label{mixed-1best} 1-best accuracy results (single run) for all experiments with the mixed data composition using the base version of the model. Data amount $0$ denotes the out-of-the-box baseline performance compared to fine-tuning with the small (S), medium (M), or large (L) data sizes.}
\end{table*}

\begin{table*}[h]\centering\small
\begin{tabular}{@{}cc|rrrrrrrrrr@{}}
\toprule
        &        & \multicolumn{10}{c}{\textbf{Interventions}}                                                                                                                                                                               \\ \midrule
\textbf{Model} & \textbf{Data} & \multicolumn{1}{c}{\textbf{None}} & \multicolumn{1}{c}{\textbf{IPA}} & \multicolumn{1}{c}{\textbf{Shift}} & \multicolumn{1}{c}{\textbf{Reg}} & \multicolumn{1}{c}{\textbf{Char}} & \multicolumn{1}{c}{\textbf{Pig}} & \multicolumn{1}{c}{\textbf{$-$End}} & \multicolumn{1}{c}{\textbf{Affix}} & \multicolumn{1}{c}{\textbf{Hyp}} & \multicolumn{1}{c}{\textbf{Ant}} \\ \midrule
BERT      & 0       & 62.0               & 0.1               & 0.0                & 12.1               & 0.0                & 0.0               & 1.6                & 0.1                & 0.0               & 0.2               \\
        & S       & 55.1               & 0.3               & 0.0                & 9.9               & 0.5                & 0.0               & 34.2                & 0.1                & 1.8               & 3.2               \\
        & M       & 51.3               & 2.4               & 2.3                & 12.5               & 1.2                & 0.2               & 30.7                & 3.2                & 10.1               & 18.1               \\
        & L       & 50.3               & 9.8               & 2.2                & 10.2               & 7.3                & 7.5               & 39.0                & 25.2                & 31.2               & 31.5               \\
\midrule mBERT     & 0       & 44.8               & 1.2               & 0.9                & 6.8               & 0.9                & 1.4               & 2.9                & 0.5                & 1.7               & 0.8               \\
        & S       & 40.7               & 1.2               & 0.4                & 1.8               & 0.8                & 0.9               & 25.5                & 0.4                & 7.0               & 10.4               \\
        & M       & 33.6               & 2.7               & 0.2                & 9.9               & 8.1                & 2.8               & 33.3                & 9.0                & 13.5               & 17.1               \\
        & L       & 37.0               & 5.3               & 0.8                & 8.5               & 8.9                & 7.7               & 32.2                & 31.0                & 27.7               & 30.5               \\ \bottomrule
\end{tabular}
\caption{\label{full-em} Exact match accuracy results (single run) for all experiments with the full data composition using the base version of the model. Data amount $0$ denotes the out-of-the-box baseline performance compared to fine-tuning with the small (S), medium (M), or large (L) data sizes.}
\end{table*}

\begin{table*}[h]\centering\small
\begin{tabular}{@{}cc|rrrrrrrrrr@{}}
\toprule
        &        & \multicolumn{10}{c}{\textbf{Interventions}}                                                                                                                                                                               \\ \midrule
\textbf{Model} & \textbf{Data} & \multicolumn{1}{c}{\textbf{None}} & \multicolumn{1}{c}{\textbf{IPA}} & \multicolumn{1}{c}{\textbf{Shift}} & \multicolumn{1}{c}{\textbf{Reg}} & \multicolumn{1}{c}{\textbf{Char}} & \multicolumn{1}{c}{\textbf{Pig}} & \multicolumn{1}{c}{\textbf{$-$End}} & \multicolumn{1}{c}{\textbf{Affix}} & \multicolumn{1}{c}{\textbf{Hyp}} & \multicolumn{1}{c}{\textbf{Ant}} \\ \midrule
BERT      & 0       & 77.0               & 10.5               & 10.0                & 13.9               & 26.4               & 23.9               & 21.5                & 2.2                & 1.2               & 0.5               \\
        & S       & 74.3               & 17.1               & 17.3                & 23.5               & 46.6               & 33.1               & 47.1                & 2.0                & 6.9               & 8.5               \\
        & M       & 74.9               & 29.3               & 28.3                & 27.9               & 55.6               & 40.2               & 46.6                & 31.7                & 23.8               & 31.0               \\
        & L       & 62.6               & 35.2               & 33.5                & 27.6               & 60.0               & 51.8               & 55.8                & 52.6                & 45.2               & 45.9               \\
\midrule mBERT     & 0       & 64.1               & 6.9               & 8.6                & 14.3               & 25.3               & 27.4               & 21.5                & 4.1                & 2.6               & 2.3               \\
        & S       & 63.5               & 21.2               & 14.3                & 15.4               & 44.9               & 32.8               & 36.9                & 5.5                & 11.2               & 16.1               \\
        & M       & 61.6               & 24.9               & 14.4                & 29.3               & 55.2               & 35.2               & 46.9                & 31.2                & 32.5               & 36.6               \\
        & L       & 49.0               & 25.2               & 19.3                & 23.7               & 62.4               & 44.7               & 43.7                & 56.8                & 44.7               & 47.7               \\ \bottomrule
\end{tabular}
\caption{\label{full-5best} 5-best accuracy results (single run) for all experiments with the full data composition using the base version of the model. Data amount $0$ denotes the out-of-the-box baseline performance compared to fine-tuning with the small (S), medium (M), or large (L) data sizes.}
\end{table*}

\end{document}